\documentclass{article}
\usepackage{tcolorbox}
\usepackage{amsmath}
\usepackage{amssymb}
\usepackage{xcolor}      
\usepackage{colortbl}    
\usepackage{array}       
\usepackage{booktabs,multirow}
\usepackage[table]{xcolor}
\usepackage{caption}   
\DeclareMathOperator*{\argmin}{arg\,min}

\definecolor{hessiancolor}{RGB}{204, 57, 42}    
\definecolor{ggncolor}{RGB}{79, 155, 143}       
\definecolor{bhessiancolor}{RGB}{44, 97, 194}   
\definecolor{bggncolor}{RGB}{217, 116, 89}      
\definecolor{ekfaccolor}{RGB}{228, 197, 119}    
\definecolor{kfaccolor}{RGB}{155, 106, 145}     

\definecolor{hessianbg}{RGB}{245, 235, 234}     
\definecolor{ggnbg}{RGB}{235, 243, 242}         
\definecolor{bggnbg}{RGB}{245, 238, 235}        
\definecolor{ekfacbg}{RGB}{250, 247, 238}       
\definecolor{kfacbg}{RGB}{243, 238, 242}        

\definecolor{sectionbg}{RGB}{230, 230, 240}     

\PassOptionsToPackage{numbers, compress}{natbib}
 \usepackage[dblblindworkshop, final]{neurips_2025}

\usepackage[utf8]{inputenc} 
\usepackage[T1]{fontenc}    
\usepackage[%
  colorlinks = true,
  citecolor  = hessiancolor,
  linkcolor  = hessiancolor,
  urlcolor   = blue,
  unicode,
  linktocpage,
]{hyperref}        
\usepackage{url}            
\usepackage{booktabs}       
\usepackage{amsfonts}       
\usepackage{nicefrac}       
\usepackage{microtype}      
\usepackage{xcolor}         

\title{Better Hessians Matter: Studying the Impact of Curvature Approximations in Influence Functions}
\workshoptitle{Mechanistic Interpretability}

\author{
  Steve Hong\\
  University of Cambridge\\
  \texttt{mdh58@cam.ac.uk}
  \And
  Runa Eschenhagen\\
  University of Cambridge\\
  \texttt{re393@cam.ac.uk}
  \AND
  Bruno Mlodozeniec\\
  University of Cambridge\\
  \texttt{bkm28@cam.ac.uk}
  \And
  Richard E. Turner\\
  University of Cambridge\\
  Alan Turing Institute\\
  \texttt{ret26@cam.ac.uk}
}

\begin{document}

\maketitle

\begin{abstract}
Influence functions offer a principled way to trace model predictions back to training data, but their use in deep learning is hampered by the need to invert a large, ill-conditioned Hessian matrix.
Approximations such as Generalised Gauss-Newton (GGN) and Kronecker-Factored Approximate Curvature (K-FAC) have been proposed to make influence computation tractable, yet it remains unclear how the departure from exactness impacts data attribution performance.
Critically, given the restricted regime in which influence functions are derived, it's not necessarily clear better Hessian approximations should even lead to better data attribution performance.
In this paper, we investigate the effect of Hessian approximation quality on influence-function attributions in a controlled classification setting. 
Our experiments show that \textit{better Hessian approximations consistently yield better influence score quality}, offering justification for recent research efforts towards that end.
We further decompose the approximation steps for recent Hessian approximation methods and evaluate each step’s influence on attribution accuracy. 
Notably, the mismatch between K-FAC eigenvalues and GGN/EK-FAC eigenvalues accounts for the majority of the error and influence loss. These findings highlight which approximations are most critical, guiding future efforts to balance computational tractability and attribution accuracy.
\end{abstract}

\section{Introduction}

When attempting to understand the behaviour of a machine learning model, a common question is: how did the training examples contribute to a given model output? Which examples contributed the most? This can also be framed counterfactually: how would the predictions change if certain training examples were removed and the model was retrained? The goal of training–data attribution (TDA) methods \cite{hammoudeh2024training} is to answer this question in a principled way.

Among these methods, influence functions \cite{Koh2017, Grosse2023, liu2021influence, jia2019towards, hampel1974influence}
provide an efficient tool by exploiting the local structure of the loss landscape around the learned parameters. The efficiency of influence functions makes them attractive: given per-sample gradients and second-order curvature information, they approximate the effect of removing a training data point \emph{without retraining}. Influence functions have been applied successfully in large–scale deep learning. For example, they have been used in 50 billion parameter Large Language Models (LLMs) to study generalisation \cite{Grosse2023}, and in scalable data attribution for diffusion models \cite{Mlodozeniec2024}.

A key practical challenge in influence function implemenetation is the Hessian bottleneck \cite{Koh2017}. Exact computation of the inverse Hessian-vector product is intractable for modern models because the Hessian is often large and ill-conditioned \cite{Grosse2023, bae2022if}. To address this, two broad approximation regimes are used. Iterative methods such as conjugate gradient \cite{shewchuk1994introduction} or LiSSA \cite{agarwal2017second} approximate inverse Hessian–vector products by using an iterative solver. These methods are asymptotically exact, but often require thousands of steps for decent performance \cite{Koh2017}. Structured approximations, on the other hand, replace the Hessian with stable and light-weight alternatives: the Generalised Gauss–Newton (GGN) \cite{martens2020new}, block–diagonal forms \cite{pmlr-v70-botev17a}, and Kronecker–factored variants such as Kronecker-Factored Approximate Curvature (K-FAC) \cite{Martens2015}, usually with a separate eigenvalue correction step (EK-FAC) \cite{George2018} to improve spectral fidelity. However, some of the approximations K-FAC and EK-FAC make are quite specific to the optimisation setting, in which they have been shown to have other desirable properties that lead to good down-stream performance, beyond the original goal of being tractable \cite{benzing2022gradient, zhang2025concurrence}.

Considerable effort in both the optimisation and data attribution communities has recently gone into developing more faithful curvature approximations \cite{George2018, dangel2021vivit, eschenhagen2023kronecker, tselepidis2020two, wang2025better}. However, it is not obvious whether such efforts are beneficial for influence estimation: influence functions may be robust to some approximation errors, while they can be substantially sensitive to certain curvature information. Clarifying when and by how much better Hessian approximations improve influence function-based attribution would help determine whether the community should continue to invest in developing higher–fidelity curvature models, or whether the gains are marginal relative to their cost.

\paragraph{Core contributions.} This work first decomposes the three approximation layers of K-FAC and examines the literature on when each approximation holds exactly versus when it introduces error. We then design controlled experiments to empirically investigate three questions:
\begin{enumerate}
    \item Does higher-fidelity Hessian approximation improve influence scores?
    \item Which approximation layer contributes most to the error, and what causes it?
    \item Which approximation error is influence fidelity most sensitive to?
\end{enumerate}

\section{Related work}

\paragraph{Fragility of influence functions.}
\citet{basu2020influence} show that influence estimates can misalign with leave-one-out retraining and are sensitive to model depth, regularisation, and query choice. However, \citet{epifano2023revisiting} attribute part of the effect to evaluation design and claim that regularisation alone is insufficient.
Similarly, \citet{mlodozeniec2025distributional} show that leave-one-out error is in large part attributable to stochastic initialisation and training, and suggest alternative evaluation and influence formulations that take that into account. \citet{bae2022if} isolates warm starts, damping/proximity, non-convergence, linearisation, and solver terms, and argues that practical estimates often resemble a Proximal Bregman response function; solver-induced error remains underexplored. Group deletions show high rank correlation but possibly large absolute errors, clarifying when correlation is informative \citep{koh2019accuracy}. \citet{ye2025towards} propose an alternative influence function formulation that leverages flat validation minima to improve robustness. Recent work studies the LiSSA and EK-FAC approximation error with a focus on mislabel detection \citep{zhu2025revisiting}. To our knowledge, the relationship between curvature-approximation error (Hessian$\!\to$GGN, block-diagonal, K-FAC/EK-FAC) and attribution quality across training regimes, depths, and widths has not been quantified. We provide a systematic evaluation in this work.

\paragraph{Hessian approximations for influence functions.}
Early implementations used iterative IHVP solvers, notably LiSSA \citep{Koh2017, agarwal2017second}. At larger scales, EK-FAC has been used to make influence estimates tractable \citep{Grosse2023, George2018}. Two directions follow: faster and more stable iterative solvers, and higher-fidelity structured curvature (e.g., GGN/K-FAC variants). ASTRA \citep{wang2025better} combines EK-FAC preconditioning with stochastic Neumann iterations to approximate damped-GGN iHVPs; relative to block-diagonal EK-FAC estimators it reduces iterations and improves attribution accuracy across architectures. These results motivate our controlled study of how GGN substitution, block-diagonality, and Kronecker factorisation trade off computational cost and attribution fidelity.

\section{Background}

We first establish the mathematical framework for influence functions and then detail the approximation layers that make them computationally tractable.

\subsection{Data attribution with influence functions}

Consider a dataset $\mathcal{D} = \{z_i\}_{i=1}^N$ where each $z_i = (x_i, y_i)$ represents an input–output pair in supervised learning;
here, $x_i \in \mathbb{R}^{d_x}$ and $y_i \in \mathbb{R}^{d_y}$.
We fit parameters $\theta^\star \in \mathbb{R}^D$ by minimising the empirical risk:
\begin{equation}
\theta^\star \;:=\; \mathop{\argmin}\limits_{\theta\in\mathbb{R}^D} J(\theta)
\;=\; \mathop{\argmin}\limits_{\theta\in\mathbb{R}^D}\; \frac{1}{N}\sum_{i=1}^N L(z_i,\theta).
\end{equation}
We evaluate model behaviour at a query $z_q$ with a measurement $m(z_q,\theta)$ (e.g., a loss or score). For a training point $z_m\in\mathcal{D}$, an attribution method $\tau(z_q,z_m,\mathcal{D})$ quantifies how $z_m$ affects $m(z_q,\theta^\star)$. To study this effect, introduce a scalar $\epsilon$ that up- or down-weights $z_m$ and define the response function
\begin{equation}
r(\epsilon)\;:=\;\arg\min_{\theta\in\mathbb{R}^D}\;J(\theta)+\tfrac{\epsilon}{N}\,L(z_m,\theta),
\end{equation}
with $\theta^\star:=r(0)$ and $\mathbf{H}:=\nabla_\theta^2 J(\theta^\star)$.

The associated first-order stationarity condition along the path $\epsilon\mapsto r(\epsilon)$ is
\begin{equation}
0 \;=\; \nabla_\theta J(r(\epsilon)) \;+\; \tfrac{\epsilon}{N}\,\nabla_\theta L(z_m,r(\epsilon)).
\end{equation}
Differentiating this identity with respect to $\epsilon$ and evaluating at $(\theta^\star,0)$ yields
\begin{equation}
\left.\frac{dr}{d\epsilon}\right|_{\epsilon=0}
= -\,\mathbf{H}^{-1}\,\frac{1}{N}\,\nabla_\theta L(z_m,\theta^\star),
\qquad
r(\epsilon)\;\approx\;\theta^\star - \epsilon\,\mathbf{H}^{-1}\,\frac{1}{N}\,\nabla_\theta L(z_m,\theta^\star).
\end{equation}

Setting $\epsilon=-1$ corresponds to removing $z_m$ from the objective and gives the first-order parameter change
\begin{equation}
\theta^\star(\mathcal{D}\setminus\{z_m\})-\theta^\star
\;\approx\; \frac{1}{N}\,\mathbf{H}^{-1}\,\nabla_\theta L(z_m,\theta^\star).
\end{equation}

Applying the chain rule to the query metric then yields the classical influence function
\begin{equation}
\tau_{\mathrm{IF}}(z_q,z_m,\mathcal{D})
\;:=\;
\nabla_\theta m(z_q,\theta^\star)^{\!\top}\,
\mathbf{H}^{-1}\,
\nabla_\theta L(z_m,\theta^\star).
\label{eq: influence function}
\end{equation}
This provides a proxy for full retraining using only gradients at $\theta^\star$ and inverse Hessian–vector products.

\subsection{Three approximation layers of K-FAC for influence estimation}
\label{section: assumptions}

To make influence computation tractable at scale, K-FAC \cite{Martens2015} and EK-FAC \cite{George2018} are the key structured methods we use to approximate the Hessian that appears in Equation \ref{eq: influence function}. This section states the equations we evaluate and decomposes the approximation into three layers: (i) Implicit model linearisation, (ii) block-diagonal approximation, and (iii) Pre-post activation approximation (with and without eigenvalue correction).

\subsubsection{Implicit model linearisation}

The first step uses the Generalised Gauss–Newton (GGN) matrix \citep{martens2020new} as a positive-semidefinite curvature proxy for the full Hessian that removes (linearises) network curvature and only focuses on output-space curvature. The substitution avoids unstable second-derivative terms and aims to keep inversion operations well-conditioned.

\textbf{Formulation.} Let $u_i(\theta)=f(x_i;\theta)\in\mathbb{R}^{d_y}$, $\mathbf{J}_i(\theta)=\nabla_\theta u_i(\theta)$, $g_i(\theta)=\nabla_u \phi(u_i(\theta),y_i)$, and $\mathbf{H}^{(u)}_i(\theta)=\nabla_u^2 \phi(u_i(\theta),y_i)$. The empirical Hessian admits
\begin{equation*}
\mathbf{H}(\theta)
=\frac{1}{N}\sum_{i=1}^N \mathbf{J}_i^\top \mathbf{H}^{(u)}_i \mathbf{J}_i
\;+\;\frac{1}{N}\sum_{i=1}^N \sum_{k=1}^{d_y} [g_i]_k\,\nabla_\theta^2 u_{i,k}(\theta)
\;=\; \mathbf{G}(\theta) \;+\; \mathbf{R}(\theta),
\end{equation*}
where $\mathbf{G}$ is the GGN term and $\mathbf{R}$ is the residual collecting second-order parameter non-linearities. For exponential-family likelihoods, $\mathbf{G}$ coincides with the Fisher information matrix \cite{martens2020new, schraudolph2002fast}.

\paragraph{Near-optimal convergence.} First, when parameters $\theta$ yield near-optimal predictions for all training examples, the gradient of the loss with respect to outputs vanishes: $g_i(\theta) \approx 0$ for all $i$. In such cases, the residual $\mathbf{R}(\theta)$ becomes negligible irrespective of the model's intrinsic curvature, yielding $\mathbf{H}(\theta) \approx \mathbf{G}(\theta)$. This condition arises specifically even at local minima where $\nabla_\theta \mathcal{J}(\theta) = 0$. 


\paragraph{Piecewise-linear activations.} Second, for networks with piecewise-linear activation functions (e.g., ReLU), the output-curvature term $\mathbf{R}(\theta)$ is \textit{block-hollow} \citep{singh2021analytic, singh2023hessian, dangel2020modular}: within-block second derivatives (e.g., within a layer) vanish, while mixed cross-block derivatives can be non-zero \footnote{Consider a two-layer ReLU network $u = w_2 \mathrm{ReLU}(w_1 x)$, $\partial^2 u / \partial w_1^2 = \partial^2 u / \partial w_2^2 = 0$., but $\partial^2 u / \partial w_1 \partial w_2 = \mathrm{ReLU}'(w_1 x) \, x \neq 0$. This illustrates the block-hollow $\mathbf{R}(\theta)$.}. Consequently, the block-diagonal of $\mathbf{H}(\theta)$ equals the block-diagonal of $\mathbf{G}(\theta)$, while the full matrices need not be equal.

\paragraph{Neural Tangent Kernel regime.} Third, under the Neural Tangent Kernel (NTK) regime \citep{jacot2018neuralntk}, where network widths are large relative to data complexity, the model output $u_i(\theta)$ remains closely approximated by its first-order Taylor expansion around initial parameters $\theta_0$ throughout optimisation \citep{lee2019wide}. This local linearity implies $\nabla_\theta^2 u_{i,k}(\theta) \approx \mathbf{0}$ along the optimisation trajectory, rendering $\mathbf{R}(\theta)$ negligible and ensuring $\mathbf{H}(\theta) \approx \mathbf{G}(\theta)$ during training. This is subject to some assumptions in initialisation of parameters and a sufficiently small learning rate.

\paragraph{Remark.} One important remark is that the residual $\mathbf{R}(\theta)$ is not necessarily positive semi-definite and may contribute both positive and negative curvature to $\mathbf{H}(\theta)$ \citep{martens2020new}. Discarding $\mathbf{R}(\theta)$ thus removes potentially useful curvature information beyond merely suppressing instability. The optimisation literature often prioritises $\mathbf{G}(\theta)$ due to its numerical stability and since the curvature coming from output curvature $\mathbf{H_i}^{(u)}(\theta)$ is more important than those coming from $\nabla^2_\theta u_{i,k}$ in $\mathbf{R}(\theta)$ over the training trajectory. This preference does not imply that $\mathbf{R}(\theta)$ is universally irrelevant in contexts requiring full Hessian fidelity, such as in influence functions.

\subsubsection{Block-diagonal approximation}
\label{section: block diagonal}

This second approximation step masks cross-layer curvature and focuses on layer-wise (or group-wise) blocks so that inversion decouples across blocks, encouraging parallelism and memory efficiency.

\textbf{Formulation.} Partition parameters as $\theta=(\theta_1,\ldots,\theta_L)$ and approximate
\begin{equation*}
\mathbf{G}(\theta)\;\approx\;\mathrm{diag}(\mathbf{G}_1,\ldots,\mathbf{G}_L),
\qquad
\mathbf{G}(\theta)^{-1}\;\approx\;\mathrm{diag}(\mathbf{G}_1^{-1},\ldots,\mathbf{G}_L^{-1}).
\end{equation*}

\paragraph{Computational advantages.} Block-diagonal curvature is widely used \cite{Martens2015, gupta2018shampoo} because the inverse of a block-diagonal matrix decomposes into the inverses of its blocks: if $\mathbf{G}=\mathrm{diag}(\mathbf{G}_1,\ldots,\mathbf{G}_L)$ then $\mathbf{G}^{-1}=\mathrm{diag}(\mathbf{G}_1^{-1},\ldots,\mathbf{G}_L^{-1})$. This structural property enables parallel computation of each block's inverse, dramatically reducing computational cost from $\mathcal{O}(D^3)$ to $\mathcal{O}(\sum_i d_i^3)$ where $d_i$ is the dimension of block $i$. In the optimisation literature, studies of block‐diagonal methods demonstrate that simply ignoring off‐block cross‐terms can even yield superior convergence and generalisation compared to both full GGN and first‐order optimisers, while requiring substantially less memory than full‐matrix approaches \citep{zhang2017block}.

\paragraph{Cross-layer coupling interpretation.} To make precise what is discarded when one retains only the diagonal blocks, partition parameters by groups (e.g., layers) $\theta=(\theta_1,\dots,\theta_L)$ and write the output Jacobian as $\mathbf{J}=[\,\mathbf{J}_1~\cdots~\mathbf{J}_L\,]$ with $\mathbf{J}_i:=\partial u/\partial \theta_i
\in\mathbb{R}^{d_y\times n_i}$. For a convex-in-output loss with per-example output Hessian $\mathbf{H}^{(u)}$, the GGN is $\mathbf{G}=\frac{1}{N}\sum_{i=1}^N \mathbf{J}^\top \mathbf{H}^{(u)} \mathbf{J}$, which decomposes into a block matrix $\mathbf{G}=\bigl[\mathbf{G}_{ij}\bigr]_{i,j=1}^L$ with cross–block couplings
\begin{equation*}  
\mathbf{G}_{ij} \;=\; \frac{1}{N}\sum_{k=1}^N \mathbf{J}_i(x_k)^\top\, \mathbf{H}_k^{(u)}\, \mathbf{J}_j(x_k) \quad (i\neq j).
\end{equation*}

These off–diagonal terms quantify, in an $\mathbf{H}^{(u)}$–weighted inner product, how similarly two parameter blocks move the model's outputs: under squared loss, $\mathbf{H}^{(u)}=\mathbf{I}$ and $\mathbf{G}_{ij}$ reduces to the Gram overlap $N^{-1}\sum_k \mathbf{J}_i(x_k)^\top \mathbf{J}_j(x_k)$, so cross–block magnitude is driven by the alignment of the two blocks' output–sensitivities.

\paragraph{Justification for block-diagonality.} Classical analyses of one–hidden–layer MLPs reuse the same result to justify the block-diagonal structure of Hessian: \cite{collobert2004large} derives explicit off–diagonal formulas and shows that with a cross–entropy (CE) loss the factors $P_\theta(y|x)\bigl(1-P_\theta(y|x)\bigr)$ multiply those couplings, pushing them toward zero during training and yielding an (approximately) block–diagonal Hessian across units, and by contrast, with mean–squared error (MSE) the same cancellation need not occur, so off–diagonals generally persist \citep{dong2025towards}. More recently, a finite–sample–to–asymptotic theory at random initialisation proves that in linear models and in one–hidden–layer networks (under both MSE and CE) the \emph{ratio} of off–diagonal to diagonal block norms vanishes as the number of outputs/classes $C$ grows (with rates depending on the block), providing a justification for block–diagonal curvature when $C$ is large, as in modern LLMs \citep{tao2024scaling}. 

\paragraph{Remark.} These findings, however, are based solely on experiments with one-hidden-layer MLPs, their extension to deeper architectures requires further investigation. For deeper networks the block–diagonal assumption remains an approximation whose accuracy depends on how orthogonal (in the $\mathbf{H}^{(u)}$–metric) the per–block output Jacobians become in practice, a question we will probe empirically in the next chapter, particularly in the context of influence functions where curvature information might be important.

\subsubsection{Pre-post activation approximation}

The last approximation, K-FAC, uses a separable Kronecker structure to reduce matrix size and enable fast inversion for each curvature block.

\textbf{Formulation.} For layer $\ell$ with bias-augmented inputs $\bar{\mathbf{a}}_{\ell-1}\in\mathbb{R}^{M+1}$ and pre-activation gradients $\mathrm D\mathbf{s}_\ell\in\mathbb{R}^{P}$, K-FAC assumes $\bar{\mathbf{a}}_{\ell-1}$ and $\mathrm D\mathbf{s}_\ell$ are independent and approximates the GGN/Fisher block as
\begin{equation*}
\mathbf{G}_\ell \;\approx\; \mathbf{A}_{\ell-1}\otimes \mathbf{S}_\ell,
\qquad
\mathbf{A}_{\ell-1}:=\mathbb{E}[\bar{\mathbf{a}}_{\ell-1}\bar{\mathbf{a}}_{\ell-1}^{\top}],\quad
\mathbf{S}_\ell:=\mathbb{E}[\mathrm D\mathbf{s}_\ell \mathrm D\mathbf{s}_\ell^{\top}],
\end{equation*}
with $(\mathbf{A}_{\ell-1}\!\otimes\!\mathbf{S}_\ell)^{-1}=\mathbf{A}_{\ell-1}^{-1}\!\otimes\!\mathbf{S}_\ell^{-1}$.

\paragraph{Remark.}
By assuming independence between $\bar{\mathbf{a}}_{\ell-1}$ and $\mathrm D\mathbf{s}_\ell$, K-FAC loses the cross-covariance structure that couples activations and gradients on individual examples. As a result, it cannot represent effects such as parameters that rarely activate also rarely receiving large gradients, or input patterns that jointly induce high activations and large error signals. This missing information can be substantial in non-linear networks where the coupling between $\bar{\mathbf{a}}_{\ell-1}$ and $\mathrm D\mathbf{s}_\ell$ helps characterise local geometry. A spectral view makes the same point: the exact GGN block admits
\begin{equation}
\mathbf{G}_\ell=\mathbf{U}\,\boldsymbol{\Lambda}\,\mathbf{U}^\top,
\end{equation}
whereas K-FAC uses the Kronecker-factor surrogate
\begin{equation}
\mathbf{A}_{\ell-1}\otimes \mathbf{S}_\ell
= (\mathbf{U}_A\boldsymbol{\Lambda}_A\mathbf{U}_A^\top)\otimes(\mathbf{U}_S\boldsymbol{\Lambda}_S\mathbf{U}_S^\top)
= (\mathbf{U}_A\otimes \mathbf{U}_S)\,(\boldsymbol{\Lambda}_A\otimes \boldsymbol{\Lambda}_S)\,(\mathbf{U}_A\otimes \mathbf{U}_S)^\top.
\end{equation}
Although the update rotates into the Kronecker eigenbasis $\mathbf{U}_A\otimes \mathbf{U}_S$, its rescaling uses only products of marginal spectra $\boldsymbol{\Lambda}_A\otimes \boldsymbol{\Lambda}_S$. These products $\lambda^A_i\lambda^S_j$ are generally not the true variances of the full block along $(\mathbf{U}_A\otimes \mathbf{U}_S)$’s directions, leading to systematic curvature misestimation: marginal second moments are preserved, but cross-covariances are discarded, which distorts the spectrum of the true GGN block.

\subsubsection{Eigenvalue correction}

Instead, we can keep K-FAC’s Kronecker-factored eigenbasis while correcting the per-direction scaling to better match the empirical curvature.

\textbf{Formulation.} Write $\mathbf{A}_{\ell-1}=\mathbf{U}_A\boldsymbol{\Lambda}_A\mathbf{U}_A^\top$ and $\mathbf{S}_\ell=\mathbf{U}_S\boldsymbol{\Lambda}_S\mathbf{U}_S^\top$, and let $\mathbf{U}:=\mathbf{U}_A\otimes \mathbf{U}_S$. If $g_\ell$ denotes the (vectorised) per-layer gradient, EK-FAC sets
\begin{equation*}
s_k^\star \;:=\; \mathbb{E}\!\left[\bigl(\mathbf{U}^\top g_\ell\bigr)_k^2\right], 
\qquad
\mathbf{S}^\star:=\mathrm{diag}(s_1^\star,\dots,s_K^\star),
\qquad
\mathbf{G}_\ell \;\approx\; \mathbf{U}\,\mathbf{S}^\star\,\mathbf{U}^\top.
\end{equation*}

\textbf{Remark.}
However, EK-FAC does not correct the direction of the approximation: it retains K-FAC’s Kronecker-factored eigenbasis $\mathbf{U}_A \otimes \mathbf{U}_S$. The update only corrects per-coordinate scaling by matching the Fisher’s diagonal in that basis, so any genuine coupling between coordinates, i.e., curvature that appears as off-diagonal mass in Kronecker eigenbasis coordinates remains unmodelled. Thus, when the true block $\mathbf{G}_\ell$ has principal directions that are not well captured by a separable Kronecker structure, eigenvalue correction alone cannot recover those interactions as it rescales coordinates rather than also rotating them.

\section{Investigating the approximation error \& influence score relationship}

\subsection{Experimental setup}
\label{section: experiment setup}

\textbf{Objective.} To understand how each approximation layer impacts influence quality, we need experimental settings where the approximation errors vary systematically. In Section \ref{section: assumptions} we identified that each approximation layer---GGN substitution, block-diagonalisation, and Kronecker factorisation---introduces different error types that depend on the model's curvature properties. We therefore design experiments along three dimensions that naturally modulate these curvature characteristics: (i) training duration, where early training exhibits large residual terms that diminish near convergence; (ii) network depth, which amplifies cross-layer coupling and non-linear interactions between parameters; and (iii) network width, which affects the conditioning and spectral properties of individual layer blocks. These controlled variations allow us to isolate when each approximation breaks down and quantify its impact on attribution fidelity.

\paragraph{Dataset.} We use the Digits dataset \citep{alpaydin1998optical}, which contains $n=1{,}797$ greyscale images of handwritten digits (0--9). Each $8\times8$ image is converted into a $64$-dimensional vector. We randomly split the data into $n_{\mathrm{train}}=1{,}617$ training samples (90\%) and $n_{\mathrm{test}}=179$ test samples, maintaining equal representation of all digit classes.

\paragraph{Model architecture.} Due to computational constraints, we restrict our experiments to multi-layer perceptrons (MLPs). The specific training settings are specified in the hyperparameter settings section below, and the limitations of this choice are discussed in the Section \ref{section: limitations}. We use Tanh activation functions throughout, which ensure non-convexity and that the residual term exists, allowing us to isolate the effects of training.

\paragraph{Matrix inversion and numerical stability.}
As shown in Section~\ref{section: assumptions}, the Hessian is often ill-conditioned even for simple models. To address invertibility, one option is to add a Tikhonov damping term \(\lambda\) to the diagonal of each matrix. For this experiment, this approach biases curvature differences between methods, which makes the comparison unfair. We therefore adopt a second approach: pseudo-inverse computation. The conventional choice is the Moore--Penrose pseudo-inverse via SVD \citep{eckart1936approximation}. However, we instead use an eigendecomposition-based pseudo-inverse \citep{wright1999numerical} for two reasons: (i) the decomposition exists for square symmetric matrices, which is the case for the Hessian; (ii) we often want to regularise the matrix to be positive definite, which is simpler with eigenvalue adjustments than with SVD.

Formally, for a symmetric matrix \(\mathbf{H}=\mathbf{Q}\mathbf{\Lambda}\mathbf{Q}^\top\), the pseudo-inverse is \(\mathbf{H}^\dagger=\mathbf{Q}\mathbf{\Lambda}^\dagger\mathbf{Q}^\top\) with diagonal entries \([\mathbf{\Lambda}^\dagger]_{ii}=\lambda_i^{-1}\,\mathbf{1}\{|\lambda_i|>\epsilon\}\) and zero otherwise. We set \(\epsilon=10^{-4}\) and use no damping in this section. Sensitivity to \(\epsilon\) is potential future work and may be relevant for interpreting the results; see Section~\ref{section: limitations}.


\paragraph{Evaluation metrics.} We employ two complementary metrics to assess both the quality of influence attributions and the fidelity of Hessian approximations:

\begin{itemize}
\setlength{\leftskip}{-2em}  
    \item \textbf{Linear data-modelling score (LDS):} Following the framework described in Appendix \ref{section: LSO}, we use the expected leave-some-out evaluation with subset fraction $\alpha$.
    
    \item \textbf{Approximation error:} We cannot reliably use \( \mathbf{H}^{-1} \) as a reference because \( \mathbf{H} \) is typically singular or nearly singular in our setting; instead we assess whether \( \mathbf{H} \widehat{\mathbf{H}}^{-1} v \approx v \). For a set of vectors $\{v_i\}_{i=1}^N$ (using training data gradients), we compute:
    \begin{equation}
    \text{Approximation Error} = \frac{1}{N} \sum_{i=1}^N \frac{\|\mathbf{H} \cdot \widehat{\mathbf{H}}^{-1}v_i - v_i\|^2}{\|v_i\|^2}
    \label{eq: approx error}
    \end{equation}
\end{itemize}

\subsection{Results}
\label{section: results}

We now present our empirical findings addressing the three core questions posed in the introduction, examining the relationship between approximation fidelity and attribution quality across our controlled experimental conditions.

\noindent\textbf{(1) Does a better Hessian approximation improve influence scores?}

\begin{figure*}[!htb]
    \centering
    \begin{minipage}[t]{0.535\textwidth}
        \centering
        \vspace{0pt}
        \includegraphics[width=\linewidth,trim={0cm 0cm 0cm 0cm},clip]{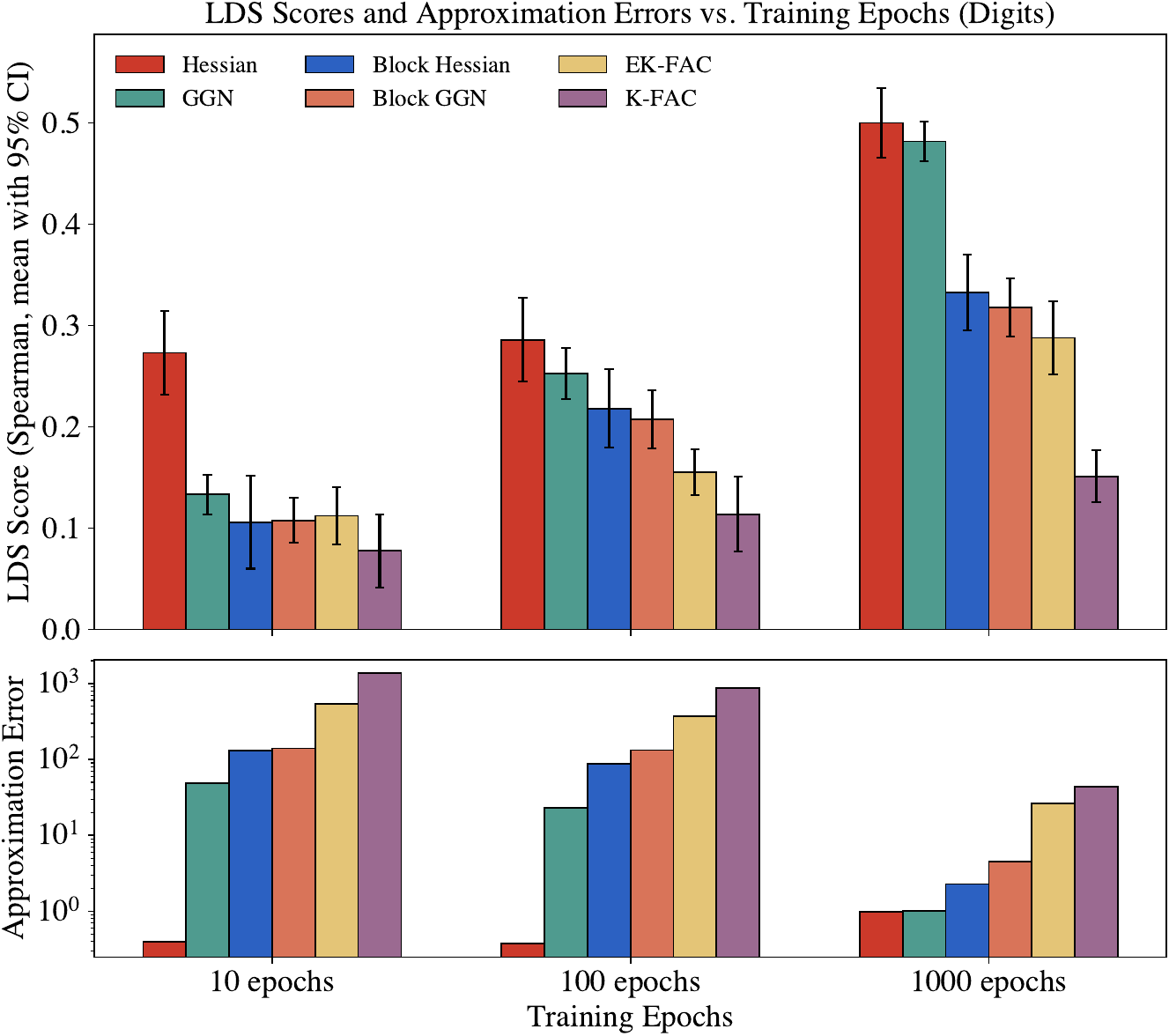}
    \end{minipage}
    \hspace{0.02\textwidth}
    \begin{minipage}[t]{0.38\textwidth}
        \centering
        \vspace{0pt}
        \footnotesize
        \begin{tabular}{l>{\columncolor{white}}r>{\columncolor{white}}r}
            \toprule
            \rowcolor{sectionbg}
            \textbf{Step (incremental)} & \textbf{$\Delta$ES$\%$} & \textbf{$\Delta$LDS$\%$} \\
            \midrule
            \rowcolor{sectionbg}
            \multicolumn{3}{c}{\textbf{Epoch = 10}} \\
            Hessian $\to$ GGN & 3.52 & \textbf{-71.59} \\
            GGN $\to$ B-GGN & 6.70 & -13.12 \\
            B-GGN $\to$ EK-FAC & 29.58 & 2.29 \\
            EK-FAC $\to$ K-FAC & \textbf{60.20} & -17.58 \\
            \midrule
            \rowcolor{sectionbg}
            \multicolumn{3}{c}{\textbf{Epoch = 100}} \\
            Hessian $\to$ GGN & 2.59 & -19.35 \\
            GGN $\to$ B-GGN & 12.55 & -26.25 \\
            B-GGN $\to$ EK-FAC & 26.80 & \textbf{-30.39} \\
            EK-FAC $\to$ K-FAC & \textbf{58.06} & -24.01 \\
            \midrule
            \rowcolor{sectionbg}
            \multicolumn{3}{c}{\textbf{Epoch = 1{,}000}} \\
            Hessian $\to$ GGN & 0.04 & -5.19 \\
            GGN $\to$ B-GGN & 8.25 & \textbf{-46.95} \\
            B-GGN $\to$ EK-FAC & \textbf{50.74} & -8.67 \\
            EK-FAC $\to$ K-FAC & 40.97 & -39.19 \\
            \bottomrule
        \end{tabular}
    \end{minipage}
    \caption{\textbf{\textit{Left:} Attribution quality vs. Hessian approximation error - Training duration.} LDS and approximation error (Equation \ref{eq: approx error}); for epoch \{10, 100, 1,000\}. Setting is fixed at depth $=$ 8 and width $=$ 16; other hyperparameters follow Table~\ref{tab:hyperparameters_digits}. \textbf{\textit{Right:} Error decomposition table}: incremental shares along the curvature-approximation path. $\Delta$ES$\%$ denotes Error Share in percentage in the Hessian$\to$K-FAC path and $\Delta$LDS$\%$ denotes the total Hessian$\to$K-FAC LDS percentage change across steps. B-GGN denotes Block-Diagonal GGN.}
    \label{fig:result_epoch}
\end{figure*}

\begin{figure*}[!htb]
    \centering
    \begin{minipage}[t]{0.535\textwidth}
        \centering
        \vspace{0pt}
        \includegraphics[width=\linewidth,trim={0cm 0cm 0cm 0cm},clip]{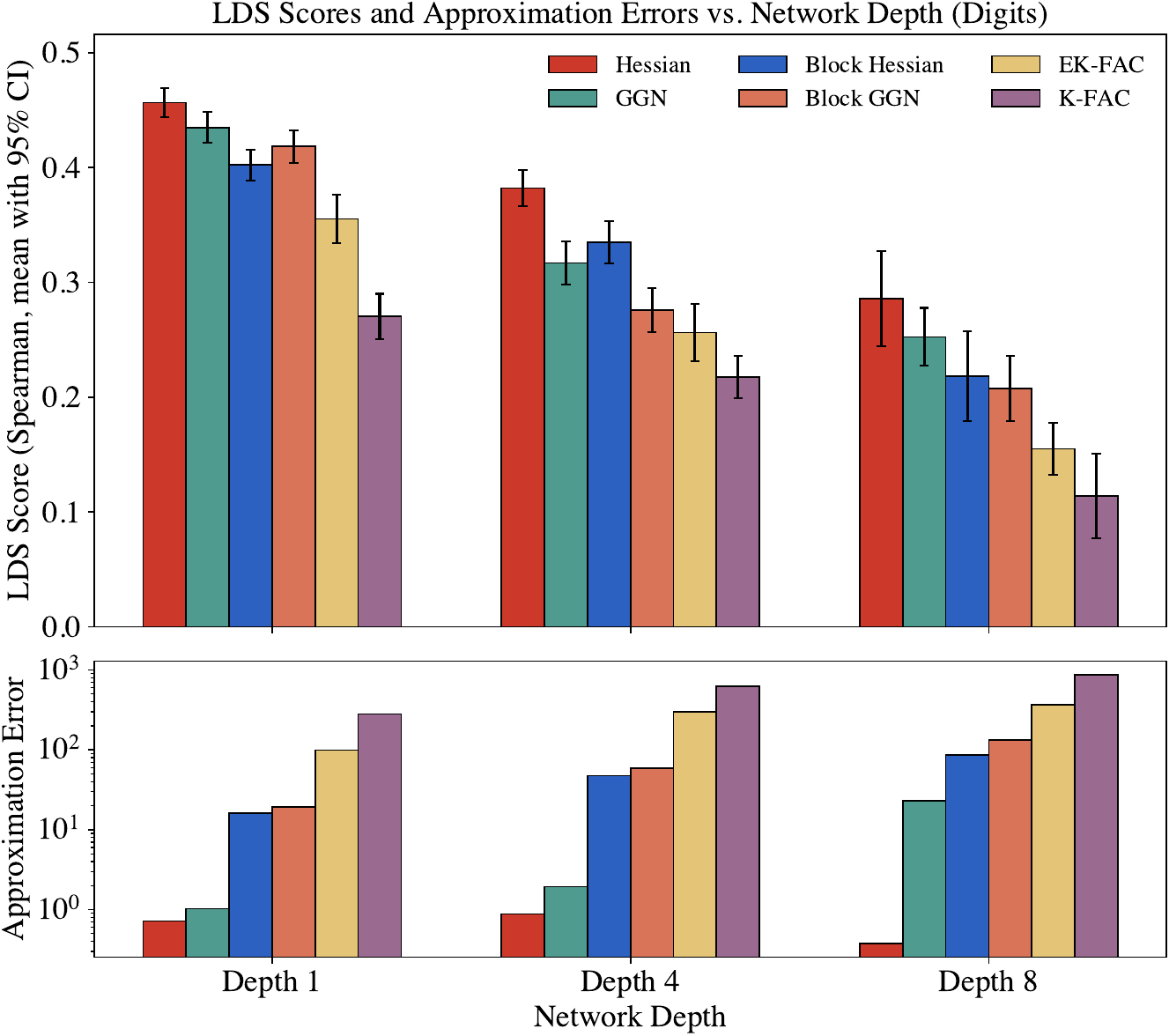}
    \end{minipage}
    \hspace{0.02\textwidth}
    \begin{minipage}[t]{0.38\textwidth}
        \centering
        \vspace{0pt}
        \footnotesize
        \footnotesize
        \begin{tabular}{l>{\columncolor{white}}r>{\columncolor{white}}r}
          \toprule
          \rowcolor{sectionbg}
          \textbf{Step (incremental)} & \textbf{$\Delta$ES$\%$} & \textbf{$\Delta$LDS$\%$} \\
          \midrule
          \rowcolor{sectionbg}
          \multicolumn{3}{c}{\textbf{Depth = 1}} \\
          Hessian $\to$ GGN & 0.11 & -11.70 \\
          GGN $\to$ B-GGN & 6.57 & -8.84 \\
          B-GGN $\to$ EK-FAC & 28.53 & -33.83 \\
          EK-FAC $\to$ K-FAC & \textbf{64.79} & \textbf{-45.63} \\
          \midrule
          \rowcolor{sectionbg}
          \multicolumn{3}{c}{\textbf{Depth = 4}} \\
          Hessian $\to$ GGN & 0.17 & \textbf{-39.73} \\
          GGN $\to$ B-GGN & 9.23 & -25.00 \\
          B-GGN $\to$ EK-FAC & 38.59 & -11.67 \\
          EK-FAC $\to$ K-FAC & \textbf{52.01} & -23.60 \\
          \midrule
          \rowcolor{sectionbg}
          \multicolumn{3}{c}{\textbf{Depth = 8}} \\
          Hessian $\to$ GGN & 2.59 & -19.35 \\
          GGN $\to$ B-GGN & 12.55 & -26.25 \\
          B-GGN $\to$ EK-FAC & 26.80 & \textbf{-30.39} \\
          EK-FAC $\to$ K-FAC & \textbf{58.06} & -24.01 \\
          \bottomrule
        \end{tabular}
    \end{minipage}
    \caption{\textbf{\textit{Left:} Attribution quality vs. Hessian approximation error - Network depth.} LDS and approximation error (Equation \ref{eq: approx error}); for depth \{1, 4, 8\}. Setting is fixed at epoch $=$ 100 and width $=$ 16; other hyperparameters follow Table~\ref{tab:hyperparameters_digits}. \textbf{\textit{Right:} Error decomposition table}: incremental shares along the curvature-approximation path. $\Delta$ES$\%$ denotes Error Share in percentage in the Hessian$\to$K-FAC path and $\Delta$LDS$\%$ denotes the total Hessian$\to$K-FAC LDS percentage change across steps. B-GGN denotes Block-Diagonal GGN.}
    \label{fig:result_depth}
\end{figure*}

\begin{figure*}[!htb]
    \centering
    \begin{minipage}[t]{0.535\textwidth}
        \centering
        \vspace{0pt}
        \includegraphics[width=\linewidth,trim={0cm 0cm 0cm 0cm},clip]{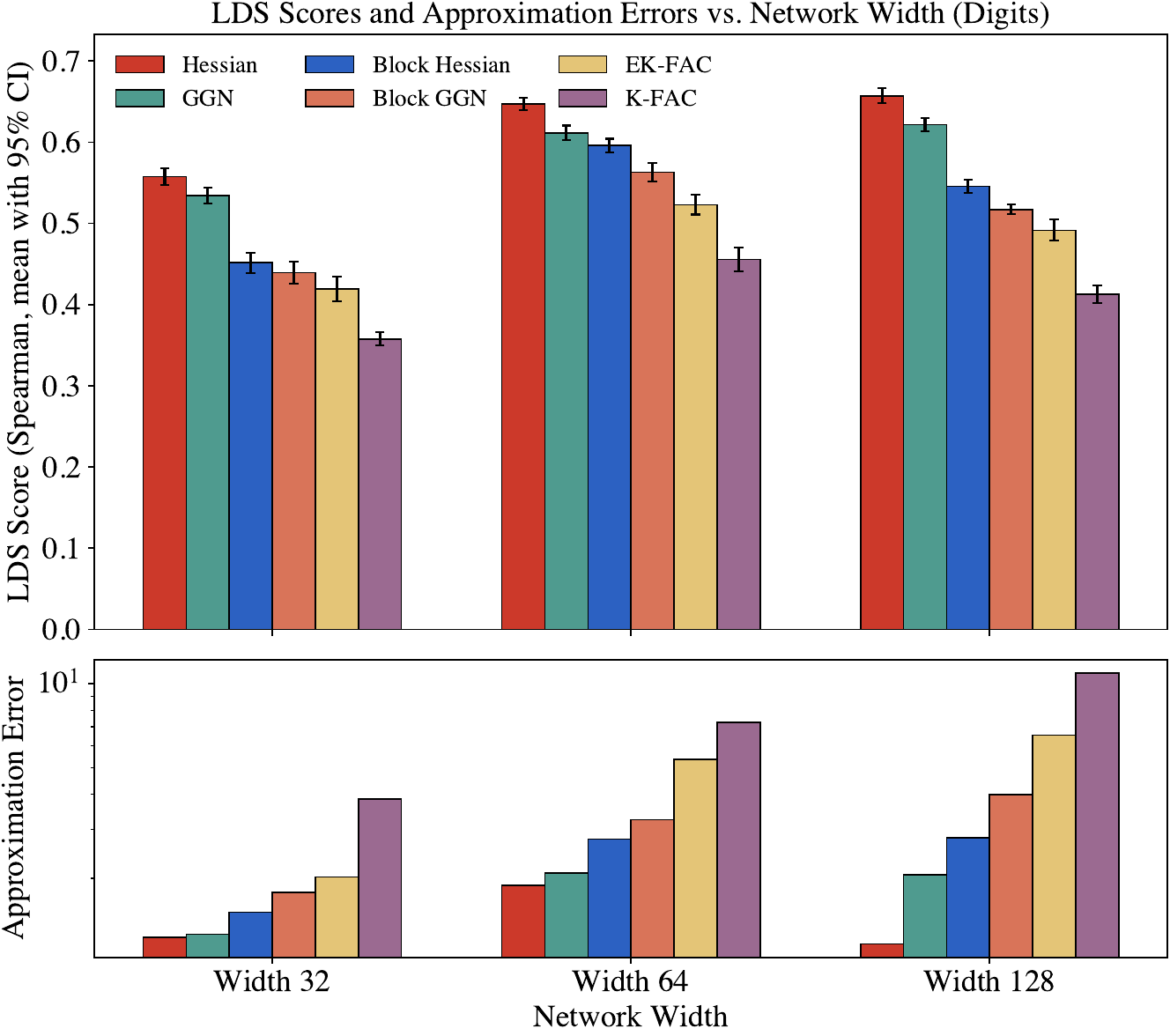}
    \end{minipage}
    \hspace{0.02\textwidth}
    \begin{minipage}[t]{0.38\textwidth}
        \centering
        \vspace{0pt}
        \footnotesize
        \begin{tabular}{l>{\columncolor{white}}r>{\columncolor{white}}r}
          \toprule
          \rowcolor{sectionbg}
          \textbf{Step (incremental)} & \textbf{$\Delta$ES$\%$} & \textbf{$\Delta$LDS$\%$} \\
          \midrule
          \rowcolor{sectionbg}
          \multicolumn{3}{c}{\textbf{Width = 32}} \\
          Hessian $\to$ GGN & 1.32 & 11.77 \\
          GGN $\to$ B-GGN & 19.88 & \textbf{47.29} \\
          B-GGN $\to$ EK-FAC & 9.11 & 10.03 \\
          EK-FAC $\to$ K-FAC & \textbf{69.69} & 30.91 \\
          \midrule
          \rowcolor{sectionbg}
          \multicolumn{3}{c}{\textbf{Width = 64}} \\
          Hessian $\to$ GGN & 3.79 & 18.57 \\
          GGN $\to$ B-GGN & 21.61 & 25.41 \\
          B-GGN $\to$ EK-FAC & \textbf{39.23} & 20.80 \\
          EK-FAC $\to$ K-FAC & 35.37 & \textbf{35.22} \\
          \midrule
          \rowcolor{sectionbg}
          \multicolumn{3}{c}{\textbf{Width = 128}} \\
          Hessian $\to$ GGN & 9.23 & 14.47 \\
          GGN $\to$ B-GGN & 19.85 & \textbf{42.82} \\
          B-GGN $\to$ EK-FAC & 26.17 & 10.48 \\
          EK-FAC $\to$ K-FAC & \textbf{44.75} & 32.23 \\
          \bottomrule
        \end{tabular}
    \end{minipage}
    \caption{\textbf{\textit{Left:} Attribution quality vs. Hessian approximation error - Network width.} LDS and approximation error (Equation \ref{eq: approx error}); for widths $\{32,64,128\}$. Setting is fixed at epoch $=$ 100 and depth $=$ 1; other hyperparameters follow Table~\ref{tab:hyperparameters_digits}. \textbf{\textit{Right:} Error decomposition table}: incremental shares along the curvature-approximation path. $\Delta$ES$\%$ denotes Error Share in percentage in the Hessian$\to$K-FAC path and $\Delta$LDS$\%$ denotes the total Hessian$\to$K-FAC LDS percentage change across steps. B-GGN denotes Block-Diagonal GGN.}
    \label{fig:result_width}
\end{figure*}

Across all settings we find a consistent inverse relationship between curvature approximation error and influence fidelity: lower error corresponds to higher LDS, with the method ordering Hessian $\gtrsim$ GGN $>$ Block-GGN $>$ EK\hbox{-}FAC $>$ K\hbox{-}FAC visible in the top (LDS) and bottom (error) panels of Figure~\ref{fig:result_epoch}–\ref{fig:result_width}. The slope of this association depends on training stage and architecture. Along training, moving from 10 to 100 to 1000 epochs tightens the cloud of method points in Figure~\ref{fig:result_epoch}: approximation error decreases while LDS increases and then saturates, and the methods cluster near convergence, indicating diminishing marginal LDS gains from additional curvature fidelity late in training. With architecture, increasing depth lowers LDS and raises approximation error for all methods (Figure~\ref{fig:result_depth}), whereas width produces smaller movements with the same ordering (Figure~\ref{fig:result_width}). 

Two diagnostics account for the stage- and architecture-dependence: cross-layer coupling (off-block mass) decreases mildly over training and increases strongly with depth (Appendix Figure~\ref{fig:error_crosslayer}), so the LDS–error slope is flatter at late epochs (weaker cross-block terms) and steeper in deeper networks (stronger cross-block terms). In addition, Kronecker spectral fidelity improves with training and worsens with depth, with EK\hbox{-}FAC showing consistently higher eigenvalue overlap than K\hbox{-}FAC (Appendix Figure~\ref{fig:error_eigenspectrum}) while the two share the same Kronecker eigenbasis and both exhibit declining basis alignment with depth (Appendix Figure~\ref{fig:error_kfe}). These properties explain why EK\hbox{-}FAC sits consistently above K\hbox{-}FAC in LDS yet remains below unfactorised Block\hbox{-}GGN, and why depth amplifies between-method gaps whereas width does not.

\noindent\textbf{(2) Which approximation layer contributes most to the error, and what caused it?}

The dominant contributor to the total Hessian$\!\rightarrow$K\hbox{-}FAC error gap is the within-block Kronecker factorisation. In the epoch sweep (Figure~\ref{fig:result_epoch}, right tables), the incremental EK\hbox{-}FAC$\!\rightarrow$K\hbox{-}FAC step accounts for $\sim\!60.2\%$ of the gap at 10 epochs, $\sim\!58.1\%$ at 100 epochs, and $\sim\!41.0\%$ at 1000 epochs. Across depth (Figure~\ref{fig:result_depth}, right), the same step remains the largest single share ($\sim\!64.8\%$ at depth~1, $\sim\!52.0\%$ at depth~4, $\sim\!58.1\%$ at depth~8). For width (Figure~\ref{fig:result_width}, right), a local exception occurs at 64 units where the Block\hbox{-}GGN$\!\rightarrow$EK\hbox{-}FAC share ($\sim\!39.2\%$) slightly exceeds EK\hbox{-}FAC$\!\rightarrow$K\hbox{-}FAC ($\sim\!35.4\%$), but taken together the two factorisation steps explain the majority of the gap at every width (about 70–78\%). 

Further diagonstic plots also clarify the mechanism. EK\hbox{-}FAC and K\hbox{-}FAC operate in the same Kronecker eigenbasis (identical basis-overlap curves; Appendix Figure~\ref{fig:error_kfe}), so moving from EK\hbox{-}FAC to K\hbox{-}FAC primarily introduces \emph{spectral mis-scaling} rather than basis error; correspondingly EK\hbox{-}FAC achieves higher eigenvalue overlap than K\hbox{-}FAC (Appendix Figure~\ref{fig:error_eigenspectrum}) but cannot close the gap to unfactorised blocks because the basis itself diverges from the true block basis as depth grows (Appendix Figure~8). The block-diagonal step (GGN$\!\rightarrow$Block\hbox{-}GGN) contributes a smaller but increasing share with depth (Figure~\ref{fig:result_depth}, right), consistent with the rise of cross-layer mass in Appendix Figure~\ref{fig:error_crosslayer}. By contrast, the GGN substitution (Hessian$\!\rightarrow$GGN) contributes little to the total error budget except early in training (Figure~\ref{fig:result_epoch}, right), which aligns with the visual compression of method differences near convergence in Figure~\ref{fig:result_epoch}.

\noindent\textbf{(3) Which approximation error is influence fidelity most sensitive to?}

Sensitivity of the relationship between approximation error and influence fidelity is also stage- and architecture-dependent. Early in training, influence fidelity is most sensitive to the Hessian$\!\rightarrow$GGN substitution: at 10 epochs a small error share ($\approx\!3.5\%$) coincides with a large LDS drop ($\approx\!-71.6$\,pp; Figure~\ref{fig:result_epoch}, right), whereas by 100 and 1000 epochs both the share and the LDS impact are much smaller (Figure~\ref{fig:result_epoch}). With increasing depth, sensitivity shifts toward block-diagonality: removing cross-block terms yields larger LDS losses per unit of error as off-block mass increases (compare, e.g., GGN$\!\rightarrow$Block\hbox{-}GGN at depth~1 vs.~8 in Figure~\ref{fig:result_depth}, right; see also Appendix Figure~\ref{fig:error_crosslayer}). Within blocks, factorisation produces the largest absolute error shares but only moderate per-share LDS penalties: EK\hbox{-}FAC’s spectral correction improves LDS relative to K\hbox{-}FAC (Figure~\ref{fig:result_epoch}–\ref{fig:result_width}), yet both share the same eigenbasis and therefore cannot recover LDS lost to basis mismatch when depth is large (Appendix \ref{fig:error_kfe}, with the associated eigenvalue trends in Appendix \ref{fig:error_eigenspectrum}). Width manipulations induce comparatively small and smooth changes; at width~64 the Block\hbox{-}GGN$\!\rightarrow$EK\hbox{-}FAC share slightly exceeds EK\hbox{-}FAC$\!\rightarrow$K\hbox{-}FAC (\ref{fig:result_width}, right), but this does not alter the qualitative ordering.

\section{Limitations}
\label{section: limitations}

\paragraph{Architecture and scale.}
The study uses a small MLP to keep the ELSO/LDS protocol tractable. This restricts depth, width, and dataset size, and narrows the curvature regimes observed. Results on absolute LDS levels and method ordering may not transfer to larger models. Very wide networks can operate closer to NTK-like regimes where the residual \(\mathbf{R}(\theta)\) between the Hessian and the GGN is smaller, potentially changing the relative benefits of linearisation, block-diagonalisation, and factorisation. In addition, the evaluation excludes CNNs and transformers, whose curvature structure differs due to weight sharing, attention, and embeddings. Replication at larger scales and on these architectures is required before drawing general conclusions.

\paragraph{Evaluation design and compute budget.}
For Digits we use \(\alpha = 0.5\), \(K = 100\) groups, and \(R = 50\) seeds (about \(5{,}000\) retrainings per setting; Table~\ref{tab:hyperparameters_digits}). This budget limits hyperparameter sweeps, the number of datasets, and repeated width–depth grids. Although ELSO reduces variance relative to leave-one-out, credible intervals remain non-negligible in early-epoch and deep settings. Larger-scale repetitions would improve precision.

\paragraph{Numerical controls and regularisation comparability.}
In Section~\ref{section: results} we compute inverse–Hessian–vector products using an eigendecomposition pseudo-inverse with a hard threshold \(\varepsilon = 10^{-4}\) and no damping (\(\lambda = 0\); Equation~\ref{section: experiment setup}). Truncation stabilises solves but introduces bias by discarding small-magnitude modes. We do not ablate \(\varepsilon\), nor compare against pure Tikhonov damping \((\mathbf{G}+\lambda\mathbf{I})^{-1}\) without truncation, so sensitivity to these controls is unknown. A systematic ablation should sweep \(\varepsilon \in [10^{-8},10^{-2}]\) and \(\lambda \in [10^{-8},10^{-1}]\) on logarithmic grids, report LDS, and log solver pathologies (non-convergence, extreme IHVP norms). Two edge cases are also informative: using only damping with no truncation, and using no damping with an extremely small truncation threshold, to separate numerical effects from approximation quality.

\section{Conclusion}
Our study provides an empirical answer to the three questions posed in the introduction. We decomposed common approximations into implicit linearisation, block-diagonal structure, and Kronecker factorisation, and evaluated attribution fidelity under expected leave-some-out retraining. Across training stages and architectural choices, better curvature fidelity generally aligned with stronger attribution, while gains narrowed near convergence. The dominant source of degradation arose from Kronecker factorisation within blocks; eigenvalue-corrected variants reduced but did not remove this gap.

\section*{Acknowledgements}
Runa Eschenhagen is supported by ARM, the Cambridge Trust, and the Qualcomm Innovation Fellowship.
Bruno Mlodozeniec is supported by the Qualcomm Innovation Fellowship.
Richard E. Turner is supported by the EPSRC Probabilistic AI Hub (EP/Y028783/1).
We thank Felix Dangel for pointing out a mistake in an earlier version of this work.

\bibliographystyle{unsrtnat}   
\bibliography{references}      

\newpage
\appendix

\section{Measuring Attribution Quality}
\label{section: LSO}

\subsection{Linear Data-modelling Score}
\label{sec: LDS}
The \emph{Linear Data-modelling Score} (LDS) provides a metric for evaluating the fidelity of training data attribution methods by simply taking the rank correlation between the ground truth scores against the predicted attribution scores. To compute the LDS, draw \(m\) random subsets 
$S_1, \dots, S_m \sim \mathcal{D}$ and for each sample/subset \(S_j\) measure the true model outcome \(f(S_j)\) (e.g., loss or accuracy after training on \(S_j\)) and the influence prediction \(\hat f(S_j)\) provided by the attribution method. The LDS is then defined as the Spearman rank correlation:
\begin{equation}
\mathrm{LDS}_{\mathcal{D}}(\hat f)
\;=\;
\rho\bigl(\{(f(S_j),\,\hat f(S_j))\}_{j=1}^m\bigr),
\label{eq:lds}
\end{equation}
capturing how faithfully \(\hat f\) predicts true model behaviour across subsets drawn from \(\mathcal{D}\).  A higher LDS thus indicates stronger predictive fidelity under the data distribution.

\subsection{Expected Leave-Some-Out Retraining}
\label{section:LSO}
The central framework being used in this work is proposed by \citep{Park2023} and has also been implemented in recent works \citep{Bae2024, wang2025better, ilyas2025magic}. It evaluate TDA methods by measuring their ability to predict the effect of removing \emph{groups} of training examples rather than individual points, the ground truth thereof can be referred to as \emph{expected leave-some-out} (ELSO) retraining. For a TDA method $\tau$ that assigns attribution scores to training examples, we leverage the additive nature of most attribution methods to compute group attributions. Given a subset $\mathcal{S} \subset \mathcal{D}$ of the training data, the group attribution for a query point $z_q$ is:
\begin{equation}
g_\tau(z_q, \mathcal{S}, \mathcal{D}; \lambda) := \sum_{z \in \mathcal{S}} \tau(z_q, z, \mathcal{D}; \lambda),
\label{eq:group_attribution}
\end{equation}
where $\lambda$ represents the hyperparameters used for training. For influence functions, this additivity follows naturally from the linearity of the first-order Taylor approximation.

\begin{figure}[!htb]
    \centering
    \includegraphics[width=\linewidth,trim={0cm 0.1cm 0cm 0.1cm},clip]{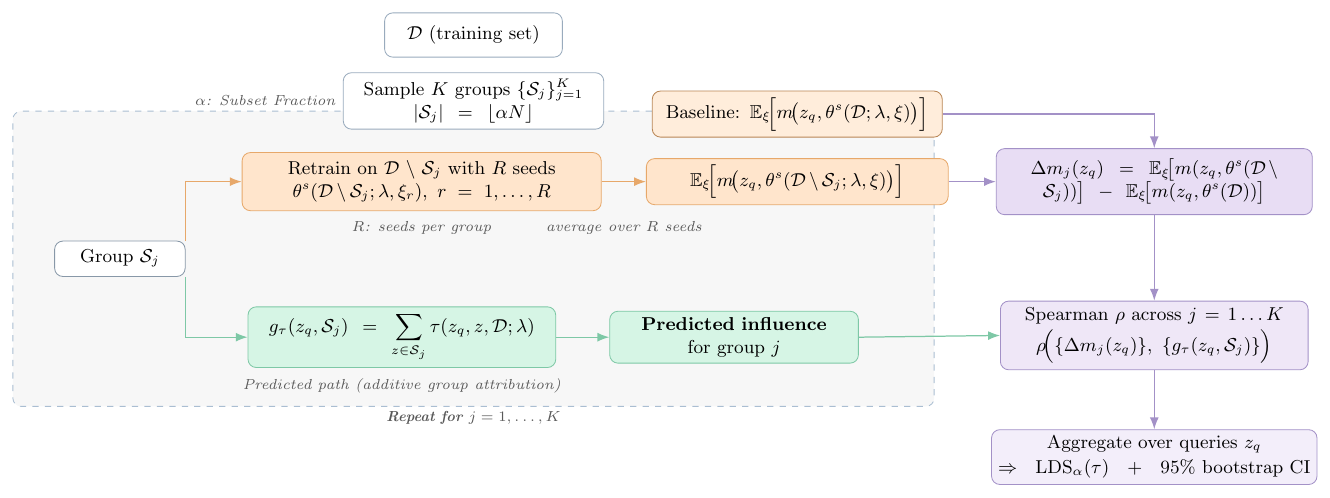}
    \caption{\textbf{Expected leave-some-out LDS evaluation.} We sample $K$ random subsets of the training data, retrain the model $R$ times per subset to average out randomness, and measure the resulting change in the query metric relative to the full-data baseline. We predict each group’s effect by summing per-example attributions, then report the Spearman rank correlation between observed and predicted effects across groups, aggregated over queries with 95\% bootstrap confidence intervals.}
    \label{fig:LDS_main}
\end{figure}

The LDS evaluation proceeds through the following systematic approach:

\begin{enumerate}
    \item \textbf{Subset generation}: We generate $K$ random subsets $\{\mathcal{S}_j\}_{j=1}^K$ from the training dataset, each containing $\lfloor \alpha N \rfloor$ data points, where $\alpha \in (0,1)$ is the data sampling ratio. This sampling ratio is crucial, too small and we lack signal, too large and we approach the computational cost of full retraining.
    
    \item \textbf{Model retraining}: For each training run with removed subset $\mathcal{S}_j$, we train the model $R$ times with different random seeds (controlling initialisation and batch ordering):
    \begin{equation}
    \{\theta^s(\mathcal{S}_j; \lambda, \xi_r)\}_{r=1}^R,
    \end{equation}
    where $\xi_r$ represents the $r$-th random seed.
    
    \item \textbf{Calculating correlation}: For a query point $z_q$, we compute the Spearman rank correlation between:
    \begin{itemize}
        \item The predicted group attributions: $\{g_\tau(z_q, \mathcal{S}_j, \mathcal{D}; \lambda) : j \in [K]\}$
        \item The actual measured effects: $\displaystyle\left\{\frac{1}{R}\sum_{r=1}^R m\bigl(z_q, \theta^s(\mathcal{S}_j; \lambda, \xi_r)\bigr) : j \in [K]\right\}$
    \end{itemize}
    
    \item \textbf{Aggregation}: The final LDS is computed by averaging correlations across multiple query points:
    \begin{equation}
    \mathrm{LDS}_\alpha(z_q, \tau) = \rho\!\Bigl(\mathbb{E}_\xi\bigl[m(z_q, \theta^s(\mathcal{S}_j; \lambda, \xi))\bigr] : j \in [K],\,\{g_\tau(z_q, \mathcal{S}_j, \mathcal{D}; \lambda) : j \in [K]\}\Bigr),
    \label{eq:lds_definition}
    \end{equation}
    where $\rho$ denotes the Spearman rank correlation.
\end{enumerate}

To ensure robust results, we report LDS scores with 95\% bootstrap confidence intervals, accounting for the randomness in subset selection.

\section{Hyperparameter Settings}
\label{section: hyperparameter}

Table~\ref{tab:hyperparameters_digits} summarises the training details for all experiments. We selected these hyperparameters through preliminary experiments to ensure models achieve reasonable convergence while maintaining computational tractability.

\begin{table}[!htb]
\centering
\small
\setlength{\tabcolsep}{8pt}
\begin{tabular}{>{\columncolor{gray!10}}l>{\columncolor{white}}l>{\columncolor{gray!10}}l>{\columncolor{white}}l}
\toprule
\rowcolor{gray!20}
\textbf{Dataset} & \textbf{Architecture} & \textbf{Training} & \textbf{ELSO Retrain} \\
\midrule
\cellcolor{gray!10}\begin{tabular}[t]{@{}l@{}}
\textbf{Digits}\\[2pt]
Train: 1,617\\
Query: 179
\end{tabular} &
\begin{tabular}[t]{@{}l@{}}
\textbf{MLP}\\[2pt]
Depth: \{1, 4, 8\}\\
Width: \{32, 64, 128\}
\end{tabular} &
\cellcolor{gray!10}\begin{tabular}[t]{@{}l@{}}
\textbf{SGD w/ Scheduler}\\[2pt]
Learning rate: 0.03\\
Weight decay: 0\\
Batch size: 32\\
Epochs: \{10, 100, 1000\}
\end{tabular} &
\begin{tabular}[t]{@{}l@{}}
\textbf{Leave-Some-Out}\\[2pt]
$\alpha$: 0.5\\
$R$: 50\\
$K$: 100\\
Total: 5,000 Models
\end{tabular} \\
\bottomrule
\end{tabular}
\vspace{0.3cm}
\caption{\textbf{Summary of training details for Digits dataset.}}
\label{tab:hyperparameters_digits}
\end{table}

For the experiments varying model architecture, we modify either the depth (1, 4, or 8 layers) or width (32, 64, or 128 hidden units per layer) while keeping other hyperparameters fixed. For the training duration experiments, we evaluate models at 10, 100, and 1000 epochs. We employ a Cosine scheduler to smoothly anneal the learning rate over training, which promotes more stable convergence. The settings for ELSO retraining follow those of \citet{Bae2024}.

\section{Attributing the Approximation Error}
\label{section:decompose_error}

We now decompose the curvature matrices into the components that most influence the approximation error quantified in the previous section. Our aim is descriptive: to identify what changes across training time, depth, and width, and to relate these changes to the approximation path introduced earlier. Throughout, \(\mathbf{H}\) denotes the exact Hessian, \(\mathbf{G}\) the GGN, and \(\mathbf{BG}\) is the exact block-diagonal GGN. For factored methods we write \(\widehat{\mathbf{G}} \in \{\text{K-FAC},\text{EK-FAC}\}\). Figures~\ref{fig:error_residual}–\ref{fig:error_kfe} summarise the measurements over the three sweeps (epochs, depth, width), and we refer back to the previous subsection for the corresponding error–LDS co-movement.

\subsection{Residual Curvature}

In order to quantify how much of the Hessian’s norm is captured by the GGN, we use the residual magnitude and track it over training epochs, network depth, and width:
\begin{equation}
\label{eq:residual_rel}
r_{\mathrm{rel}} \;=\; \frac{\lVert \mathbf{H} - \mathbf{G} \rVert_F}{\lVert \mathbf{H} \rVert_F}.
\end{equation}

\begin{figure}[!htb]
    \centering
    \includegraphics[width=0.45\textwidth]{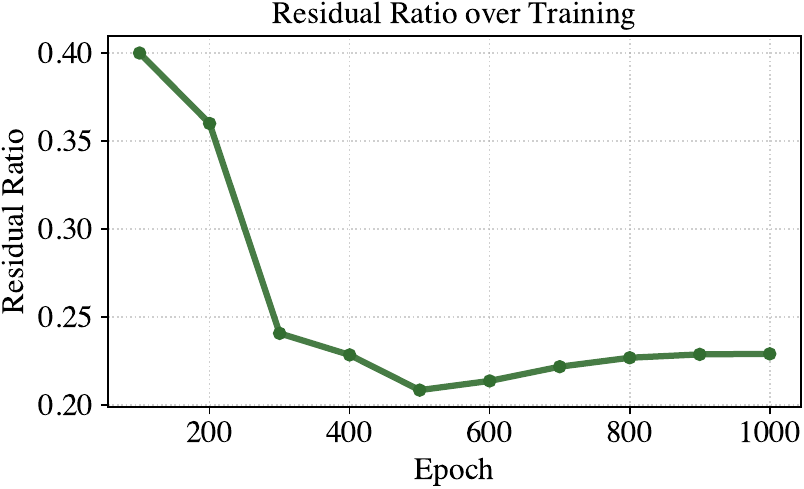}
    \hfill
    \includegraphics[width=0.45\textwidth]{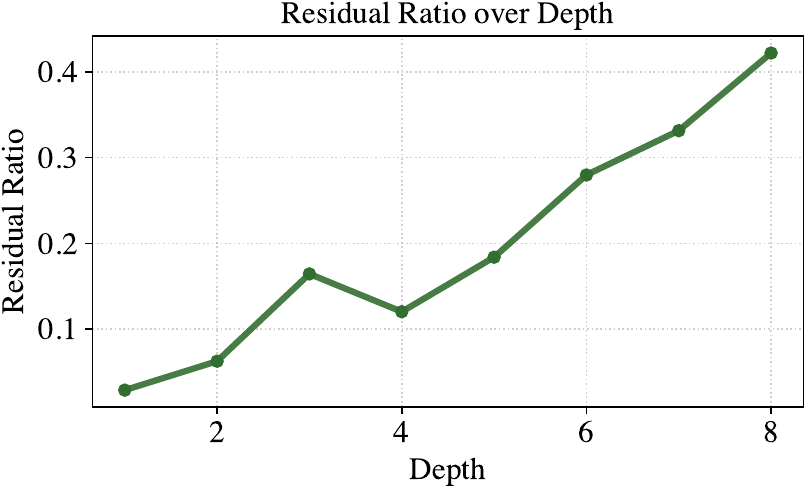}
    
    \vspace{0.3cm}
    \includegraphics[width=0.45\textwidth]{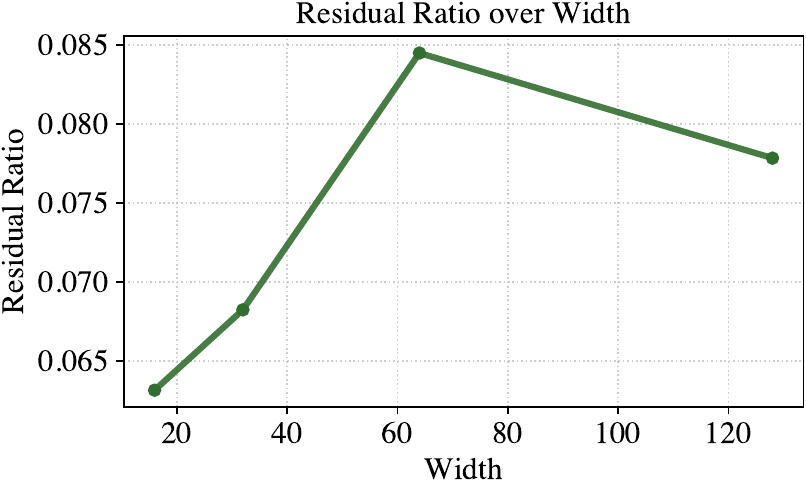}

    \caption{\textbf{Residual Term Magnitude (Digits).} Fractional size of the residual \(\mathbf{R}\) relative to the Hessian \(\mathbf{H}\) across (\textit{left}) training epochs, (\textit{right}) network depth, and (\textit{bottom}) network width. Lower values means that \(\mathbf{G}\) accounts for a larger share of \(\mathbf{H}\).}
    \label{fig:error_residual}
\end{figure}

Figure~\ref{fig:error_residual} shows a pronounced decline in the residual magnitude over the training duration, followed by a plateau at late epochs. This indicates that, as optimisation proceeds, the GGN accounts for an increasingly large fraction of the Hessian’s norm. Across network depth, the residual ratio increases from shallow to deep models, yielding a clear monotone trend. By comparison, width manipulations induce small, non-monotone changes with a much smaller dynamic range than either training time or depth.
These observations align with the linearisation perspective summarised in Section \ref{section: assumptions}: near stationary points the Taylor remainder term that separates \(\mathbf{H}\) from \(\mathbf{G}\) becomes small, whereas deeper networks, by construction, sustain a larger residual even at comparable training loss. In the context of the results in Figure \ref{fig:result_epoch}, this explains why the Hessian\,\(\to\)GGN increment contributes little to the total approximation gap at late epochs.

\subsection{Cross-Layer Curvature}

In order to measure cross-layer coupling within the GGN, we use the cross-layer curvature Frobenius norm and track it over training epochs, depth, and width:
\begin{equation}
\label{eq:offblock_ratio}
\rho_{\mathrm{cross}} \;=\; \frac{\big\lVert \mathbf{G} - \mathbf{BG} \big\rVert_F}{\lVert \mathbf{G} \rVert_F}.
\end{equation}

\begin{figure}[!htb]
    \centering
    \includegraphics[width=0.45\textwidth]{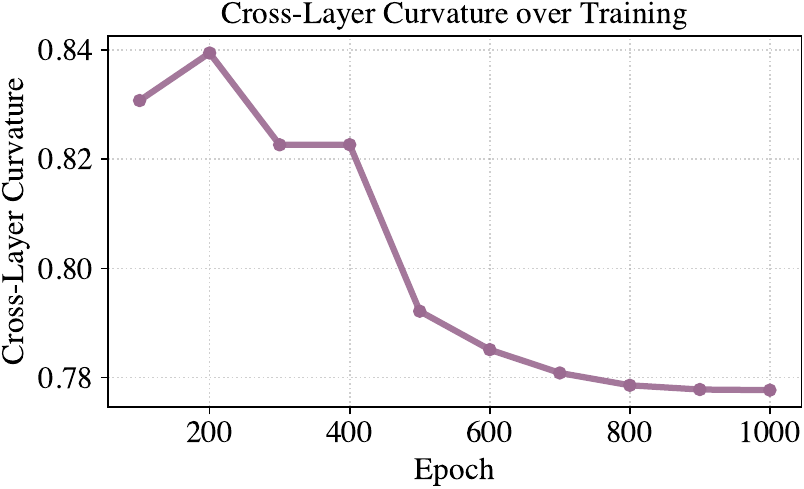}
    \hfill
    \includegraphics[width=0.45\textwidth]{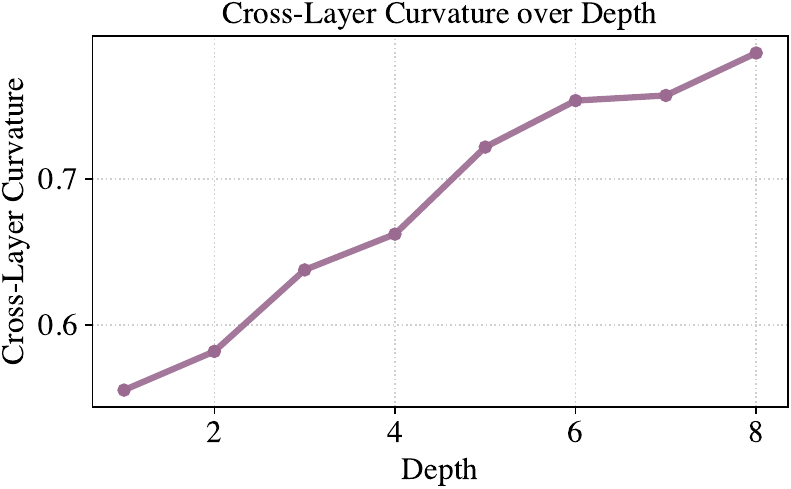}
    
    \vspace{0.3cm}
    \includegraphics[width=0.45\textwidth]{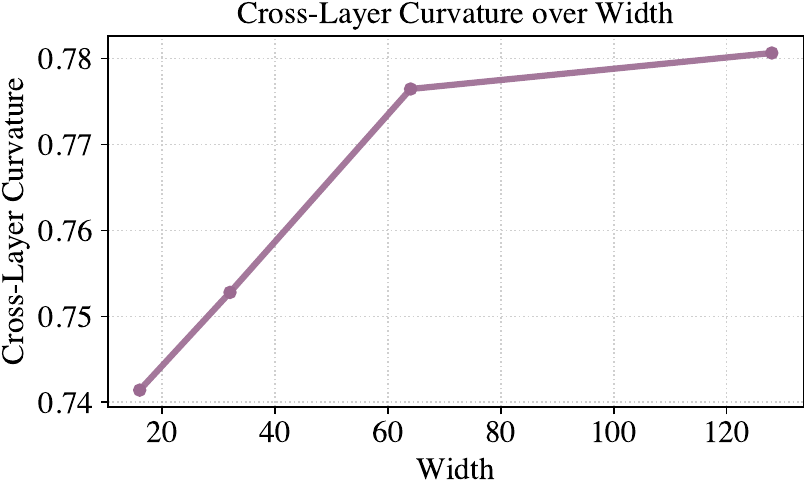}

    \caption{\textbf{Cross-layer Curvature (Digits).} Cross-layer mass of the GGN shown across (\textit{left}) training epochs, (\textit{right}) depth, and (\textit{bottom}) width. Higher values indicate stronger cross-block coupling.}
    \label{fig:error_crosslayer}
\end{figure}

The cross-layer curvature in Figure~\ref{fig:error_crosslayer} decreases slightly over training, indicating weaker cross-layer coupling as the model approaches its late-epoch operating point. In contrast, \(\rho_{\mathrm{off}}\) increases strongly with depth: deeper architectures exhibit substantially larger off-block mass in the GGN. This trend stands in contrast to the common heuristic discussed in Section \ref{section: block diagonal} that classification heads induce near block-diagonality; here, the measured cross-layer curvature expands with additional hidden layers. Width has a smaller and smoother effect: \(\rho_{\mathrm{off}}\) rises gradually with width but remains well below the magnitude changes driven by depth. The stepwise results in Figure~\ref{fig:result_depth} are consistent with these measurements: the GGN\,\(\to\)Block-GGN increment exhibits a non-negligible \(\Delta\)LDS that tracks the level of off-block mass, particularly as depth increases.

\subsection{Eigen-Spectrum Alignment}

In order to assess the spectral fidelity of Kronecker-factorised approximations, we use an eigenvalue-overlap metric between \(\widehat{\mathbf{G}}\) and the block-diagonal GGN (\(\mathbf{BG}\)), tracked across epochs, depth, and width:
\begin{equation}
\label{eq:eigen_overlap}
\mathrm{EvalOverlap}(\widehat{\mathbf{G}}, \mathbf{BG}) \;=\;
1 \;-\;
\frac{\big\lVert \operatorname{sort}\!\big(\lambda(\widehat{\mathbf{G}})\big) - \operatorname{sort}\!\big(\lambda(\mathbf{BG})\big) \big\rVert_2}
     {\big\lVert \operatorname{sort}\!\big(\lambda(\mathbf{BG})\big) \big\rVert_2}.
\end{equation}

For combining Equation \ref{eq:eigen_overlap} across all blocks, we compute each quantity per block and aggregate via a parameter count weighted average:
\begin{equation}
\label{eq:block_aggregate}
\mathrm{Agg}\;=\;\sum_{l=1}^{L} w_l \,\mathrm{Metric}_l,
\qquad
w_{l} \;=\; \frac{d_l}{\sum_{l'=1}^{L} d_{l'}},
\end{equation}
with \(d_l\) the dimensionality of block \(L\).

\begin{figure}[!htb]
    \centering
    \includegraphics[width=0.45\textwidth]{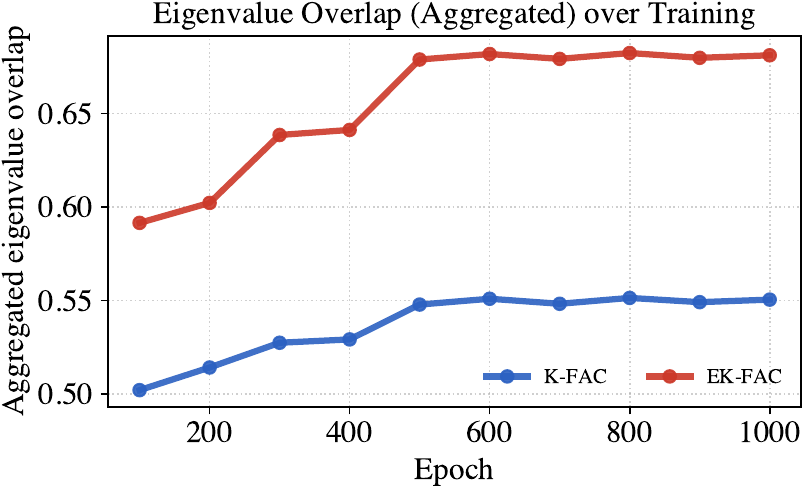}
    \hfill
    \includegraphics[width=0.45\textwidth]{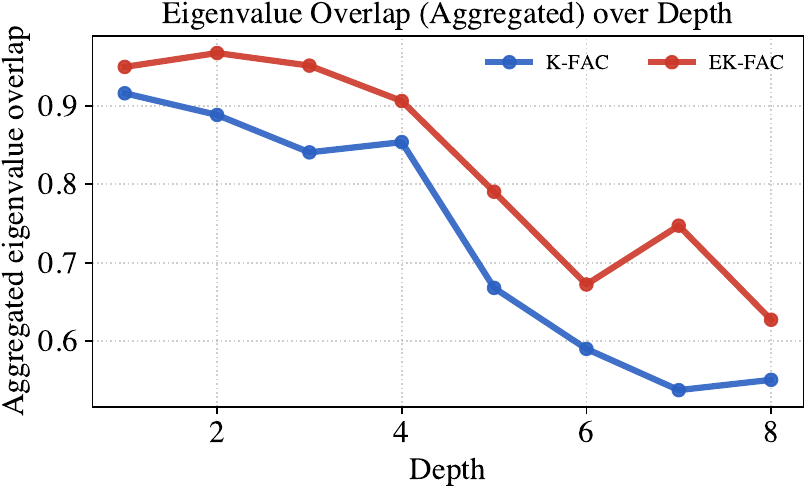}

    \vspace{0.2cm}

    \includegraphics[width=0.45\textwidth]{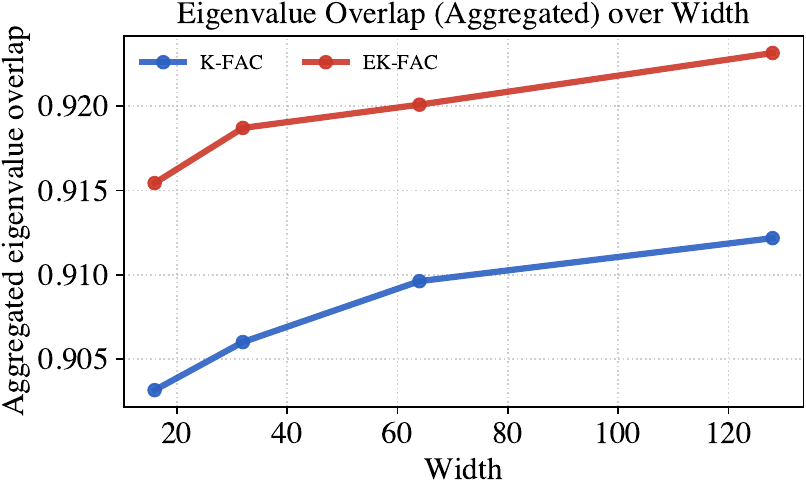}
    \caption{\textbf{Eigen-spectrum Alignment (Digits).} Aggregated eigenvalue overlap between each approximation (K-FAC, EK-FAC) and the full block GGN across (\textit{left}) training epochs, (\textit{right}) depth, and (\textit{bottom}) width. Higher values indicate closer spectral agreement.}
    \label{fig:error_eigenspectrum}
\end{figure}

Figure~\ref{fig:error_eigenspectrum} reports the aggregated overlap of eigenvalues for K-FAC and EK-FAC relative to \(\mathbf{BG}\). Over training epochs, both methods improve, with EK-FAC consistently above K-FAC and reaching a higher plateau. With increasing depth, the overlap declines for both, and the gap between EK-FAC and K-FAC widens, indicating that eigenvalue misestimation becomes more severe for the stricter Kronecker factorisation as the network deepens. Changes with width are mild and positive on average. These spectral trends mirror the recovery reported in Figure~\ref{fig:result_epoch}–\ref{fig:result_width}: EK-FAC reduces a substantial portion of K-FAC’s deficit yet does not match the full block-diagonal GGN.

\vspace{0.2cm}

\begin{figure}[!htb]
    \centering
    \includegraphics[width=0.45\textwidth]{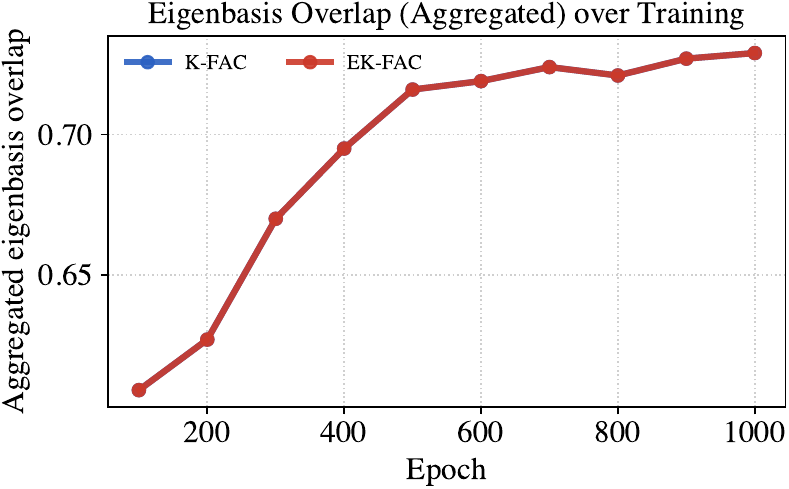}
    \hfill
    \includegraphics[width=0.45\textwidth]{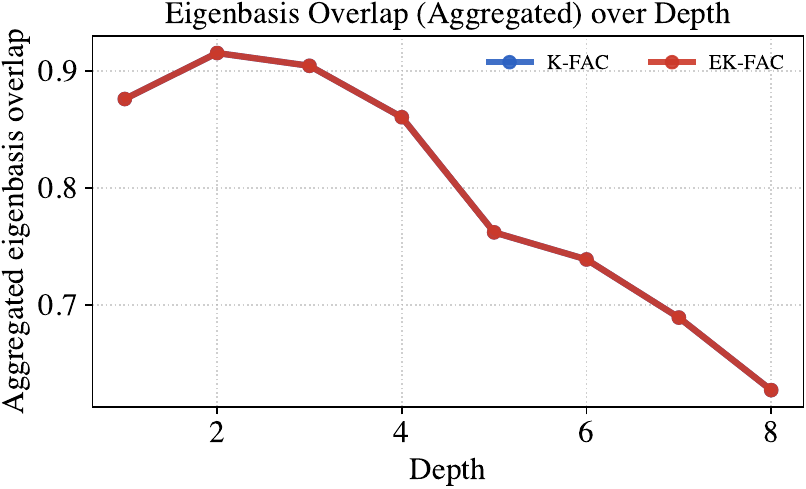}

    \vspace{0.2cm}

    \includegraphics[width=0.45\textwidth]{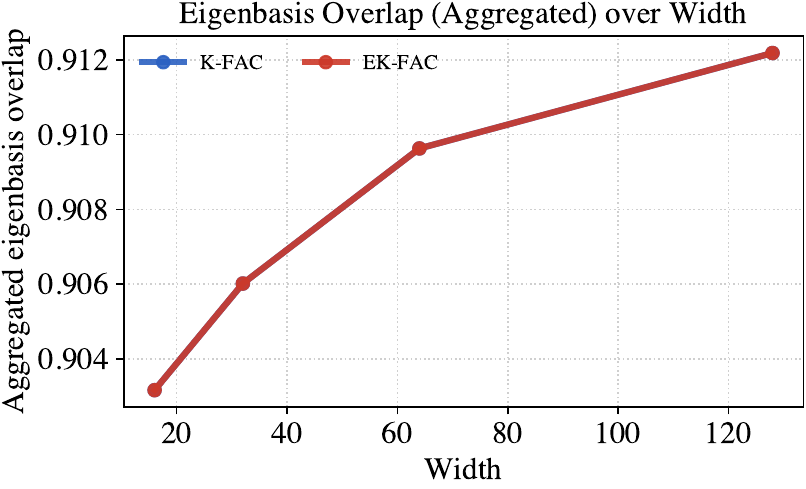}
    \caption{\textbf{Eigenbasis Alignment (Digits).} Aggregated overlap between the Kronecker-factored eigenbasis and the true eigenvectors of the full block GGN for K-FAC and EK-FAC across (\textit{left}) training epochs, (\textit{right}) depth, and (\textit{bottom}) width. Higher values indicate closer basis alignment. The two methods share the same basis, which explains the perfectly aligned plots.}
    \label{fig:error_kfe}
\end{figure}

\subsection{Kronecker-Factored Eigenbasis Alignment}

In order to evaluate Kronecker eigenbasis alignment, we use an eigenbasis-overlap metric between the eigenspaces of \(\widehat{\mathbf{G}}\) and \(\mathbf{BG}\), tracked across epochs, depth, and width:
\begin{equation}
\label{eq:basis_overlap}
\mathrm{BasisOverlap}(\widehat{\mathbf{G}}, \mathbf{BG}) \;=\;
\frac{1}{k}\,\big\lVert \mathbf{V}_k(\mathbf{BG})^{\!\top}\,\mathbf{U}_k(\widehat{\mathbf{G}})\big\rVert_F^{2}
\;=\; \frac{1}{k}\sum_{i=1}^{k}\cos^{2}\theta_i,
\end{equation}
where \(\mathbf{V}_k(\cdot)\) and \(\mathbf{U}_k(\cdot)\) collect the \(k\) eigenvectors, and \(\{\theta_i\}\) are principal angles between the corresponding subspaces. Also, $\mathbf{V}_k$ and $\mathbf{U}_k$ are chosen to be the eigenvectors in the order corresponding to the eigenvalues of each block. We choose $k$ to be the top $20\%$ of the sorted eigenvalues, which extends prior works \citep{dangel2021vivit}. Similar to Equation \ref{eq:block_aggregate}, we also use parameter count weighted average across blocks to provide a combined metric.

Figure~\ref{fig:error_kfe} shows analogous results for subspace alignment. Over training, the basis overlap increases and the two factorisations nearly coincide at late epochs. Increasing depth produces a marked decline in overlap, with K-FAC degrading more quickly than EK-FAC; this echoes the persistent dominance of the Block-GGN\,\(\to\)K-FAC increment in the total error share. Width effects are comparatively small, with EK-FAC trending flat-to-slightly-up and K-FAC showing a shallow peak at mid width followed by a modest dip.

These plots are consistent with the fact that EK-FAC operates in the same Kronecker eigenbasis as K-FAC and changes only the per direction scaling in that basis by setting the diagonal to the second moment of the projected gradient, which is the Frobenius optimal diagonal for the chosen basis; therefore both approximations share eigenvectors and exhibit identical eigenbasis overlap with the block diagonal GGN while differing primarily in eigenvalue agreement. As training proceeds the Kronecker eigenbasis tends to decorrelate gradient coordinates relative to the parameter basis, so a diagonal model in that basis becomes effective and the diagonal correction explains the observed improvement without altering the basis itself. However, as we observe that the eigenbasis overlap with the block diagonal GGN decreases as depth increases, which indicates growing basis mismatch and persistent off diagonal mass that no method constrained to be diagonal in the Kronecker factored eigenbasis can remove; consequently the approximation quality of EK-FAC worsens with depth even though its diagonal remains optimal for that fixed basis.

\end{document}